\documentclass[11pt,a4paper,table,xcdraw]{article}
\usepackage[hyperref]{emnlp2020}
\usepackage{times}
\usepackage{latexsym}

\usepackage{url}
\usepackage{hhline} 
\usepackage{graphicx}
\usepackage{CJKutf8}
\graphicspath{ {./} }
\usepackage{dblfloatfix}
\usepackage{subcaption}
\usepackage{mathtools}
\usepackage{multirow}
\usepackage[normalem]{ulem}
\useunder{\uline}{\ul}{}
\usepackage{amsmath,amsfonts,bm}
\usepackage{amssymb}%
\usepackage{pifont}%
\newcommand{\cmark}{\ding{51}}%
\newcommand{\xmark}{\ding{55}}%
\usepackage{microtype}

\aclfinalcopy %

\title{Detecting Foodborne Illness Complaints in Multiple Languages \\Using English Annotations Only}

\author{Ziyi Liu, Giannis Karamanolakis, Daniel Hsu, Luis Gravano\\
Columbia University, New York, NY 10027, USA \\
\texttt{zl2888@columbia.edu,\{gkaraman,djhsu,gravano\}@cs.columbia.edu}
}

\date{}

\begin{document}
\maketitle
\begin{abstract}
Health departments have been deploying text classification systems for the early detection of foodborne illness complaints in social media documents such as Yelp restaurant reviews.
Current systems have been successfully applied for documents in English and, as a result, a promising direction is to increase coverage and recall by considering documents in additional languages, such as Spanish or Chinese. Training previous systems for more languages, however, would be expensive, as it would require the manual annotation of many documents for each new target language.
To address this challenge, we consider cross-lingual learning and train multilingual classifiers using only the annotations for English-language reviews. 
Recent zero-shot approaches based on pre-trained multi-lingual BERT (mBERT) have been shown to effectively align languages for aspects such as sentiment. Interestingly, we show that those approaches are less effective for capturing the nuances of foodborne illness, our public health application of interest.
To improve performance without extra annotations, we create artificial training documents in the target language through machine translation and train mBERT jointly for the source (English) and target language. 
Furthermore, we show that translating labeled documents to multiple languages leads to additional performance improvements for some target languages. 
We demonstrate the benefits of our approach through extensive experiments with Yelp restaurant reviews in seven languages.
Our classifiers identify foodborne illness complaints in multilingual reviews from the Yelp Challenge dataset, which highlights the potential of our general approach for deployment in health departments. 
\end{abstract}

\section{Introduction}

\begin{figure}[h!]
\includegraphics[width=\columnwidth]{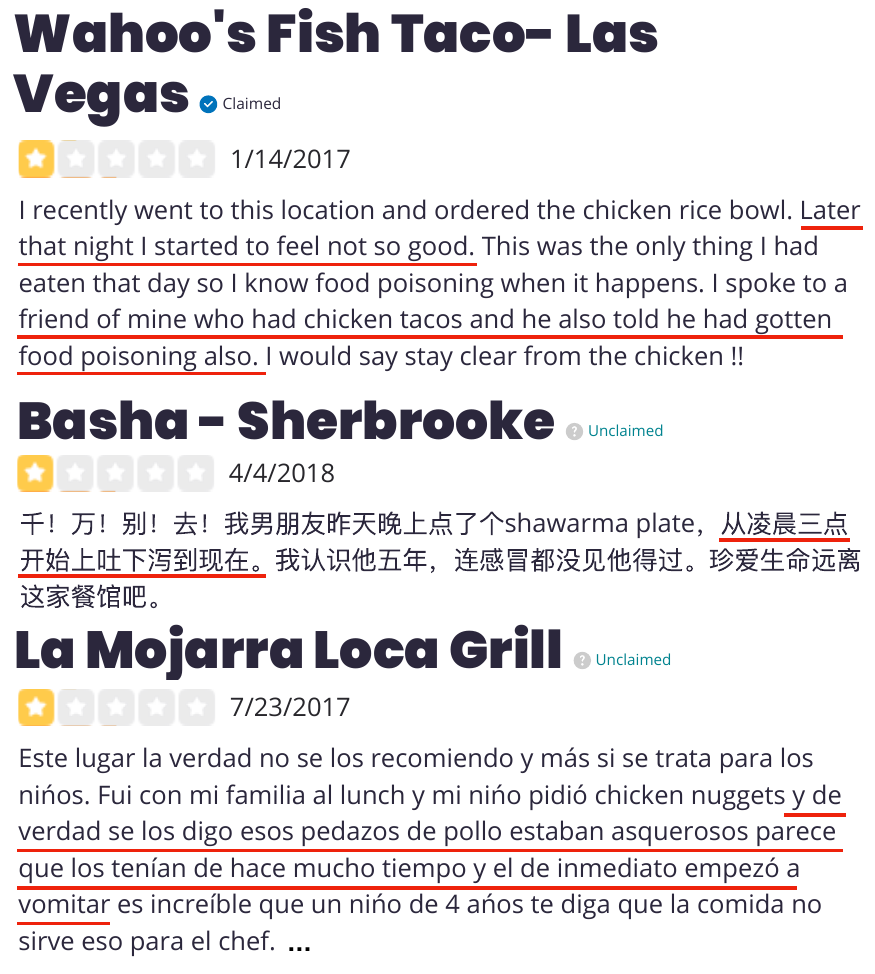}
\caption{Examples of Yelp restaurant reviews discussing food poisoning in different languages.}
\label{fig:multilingual-foodpoisoning-example}
\end{figure}
With the rise of social media, more and more users post online documents where they disclose serious incidents, such as getting food poisoning from a restaurant.  
As many of those incidents may not be reported through established complaint systems, health departments have deployed text classification systems for the identification of social media documents, such as Yelp reviews and tweets, that discuss foodborne illness episodes.  Figure~\ref{fig:multilingual-foodpoisoning-example} shows examples of Yelp restaurant reviews discussing food poisoning in English, Chinese, and Spanish. 

Current classification systems have been applied for documents written in English and deployed in several health departments, including those in Chicago~\cite{harris2014health}, Nevada~\cite{sadilek2016deploying}, New York City~\cite{effland2018discovering}, and St. Louis~\cite{harris2018evaluating}. Online documents flagged by the classifiers are typically analyzed by epidemiologists, who further investigate the incidents (e.g., by inspecting the corresponding restaurants). This process contributes to the early detection of previously unknown foodborne outbreaks. 
Given the success of current systems, a promising new direction is to extend these systems to use non-English languages, thus increasing their coverage and capacity to identify foodborne outbreaks.

Directly applying existing techniques for foodborne illness detection to other languages would be expensive and time-consuming. 
Current (supervised) classifiers have been trained on thousands of documents that were manually labeled with binary (``Sick'' vs.\ ``Not Sick'') labels provided by epidemiologists, and it would be expensive to replicate this effort for new target languages. Furthermore, it is hard to collect documents for annotation for our task because most online documents do not discuss foodborne illness. Alternative approaches beyond supervised learning are thus required to efficiently scale to multiple languages. 

To address the costly requirement of supervised learning approaches, we train multilingual classifiers through a less expensive \emph{cross-lingual} text classification approach. For a given non-English target language, our approach does not require manually annotated in-language documents but instead trains classifiers using the already available English annotations.
We follow recent techniques for cross-lingual text classification and employ pre-trained multi-lingual BERT (mBERT) representations~\cite{wu2019beto,pires2019multilingual}.
However, while pre-trained mBERT representations have been shown to be effective for tasks such as cross-lingual sentiment classification~\cite{wu2019beto}, we show that such representations are less effective for capturing the nuances of foodborne illness, which is required by our application of focus. To improve performance, we translate labeled English reviews to the target language and fine-tune mBERT \emph{jointly} for both languages, which turns out to be more effective than fine-tuning on either language separately.  
Furthermore, we show that fine-tuning mBERT for multiple languages in parallel leads to additional improvements for some target languages such as German and Italian. 

Our work makes the following contributions: 
\begin{enumerate}
    \item We present a cross-lingual learning approach for foodborne illness detection in non-English social media documents. Our approach is efficient and requires only English labeled data. 
    \item We show how to improve the performance of pre-trained mBERT for our rare classification task. 
     Our preliminary results show that generating additional artificial training data in multiple languages through machine translation leads to promising improvements over zero-shot mBERT.
    \item We evaluate our approach on Yelp reviews in English, Spanish, Chinese, French, German, Japanese, and Italian. Our approach substantially outperforms previous techniques and baselines for this task. Our multilingual classifiers successfully identify foodborne illness across languages in reviews from the Yelp Challenge dataset, which highlights the potential of our approach for successful, real-world deployment in health departments. 
\end{enumerate}
The rest of this paper is organized as follows.
In Section~\ref{s:background}, we provide the necessary background for our work. In Section~\ref{s:foodborne-model}, we describe our approach for cross-lingual foodborne detection. 
In Section~\ref{s:experiments}, we present the experimental setup and results. 
In Section~\ref{s:conclusion}, we conclude and suggest future work. 

\section{Background}
\label{s:background}
In this section, we provide background on foodborne illness detection (Section~\ref{s:background-foodborne-illness-detection}) and cross-lingual text classification (Section~\ref{s:background-crosslingual-text-classification}). 

\subsection{Foodborne Illness Detection in English Documents}
\label{s:background-foodborne-illness-detection}
Foodborne illness detection in online documents has been addressed as a binary text classification task: the goal is to train a classifier that, given the text of a document, predicts a binary (``Sick'' vs. ``Not Sick'') label, corresponding to whether the document is mentioning foodborne illness or not. 
  \citet{sadilek2016deploying} trained support vector machine classifiers (based on unigram, bigram and trigram features) using 8,000 tweets that were independently labeled by five human annotators.
 \citet{effland2018discovering} trained classifiers using more than 10,000 Yelp reviews that were manually annotated by epidemiologists.
 The paper compares several methods and found that logistic regression had the best performance.
 \citet{karamanolakis2019weakly} trained a weakly-supervised neural network that predicts a label for each individual sentence of a review and improves the recall of foodborne illness complaints compared to the best performing classifier in~\citet{effland2018discovering}. 
\subsection{Cross-Lingual Text Classification}
\label{s:background-crosslingual-text-classification}
Cross-lingual text classification trains a classifier on a target language $T$ by leveraging labeled documents in a source language $S$. 
We focus on the challenging cross-lingual classification setting where only unlabeled documents are available in $T$. 

Some effective approaches address cross-lingual classification by relying on cross-lingual word embeddings~\cite{gouws2015simple, ruder2019survey}, which represent words from different languages in the same vector space, where words across languages with similar meanings are represented as similar vectors. 
Cross-lingual word embeddings facilitate cross-lingual model transfer as a classifier trained on labeled documents in $S$ could be directly applied for test documents in $T$. 

More recent approaches addressed cross-lingual transfer using Multilingual BERT~\cite{wu2019beto,pires2019multilingual,wang2019cross,rogers2020primer}. 
Multilingual BERT, or mBERT, is a version of BERT~\cite{Devlin:19} that was trained on 104 languages in parallel. 
Training mBERT on English documents was shown to achieve impressively high performance on different target languages for several document classification tasks such as sentiment classification or topic detection~\cite{rogers2020primer}.
The successful application of mBERT for various cross-lingual tasks inspired us to employ mBERT for our public-health application, as we describe next.  

\section{Foodborne Illness Detection in Multiple Languages}
\label{s:foodborne-model}
We now define our problem of focus (Section~\ref{s:problem-definition}) and describe our cross-lingual learning approach (Sections~\ref{s:model-finetuning-on-S-T} and~\ref{s:multiple-source-language}).

\subsection{Problem Definition}
\label{s:problem-definition}
Our goal is to address foodborne illness detection in non-English languages where labeled documents are not available.
As the collection of manual annotations for each new language is an expensive and time-consuming proposition, we focus on training multilingual classifiers using only already available English documents. 
More formally, we assume access to a source language $S$ (English) with a labeled dataset $D_S = \{(x_i^S, y_i^S)\}$, where $x_i^S$ is a source language document and $y_i^S$ is the corresponding binary (``Sick'' vs. ``Not Sick'') label. For a target language $T$ we assume access to a dataset $D_T$ of unlabeled target documents $x^T$.
Our goal is to train a classifier for the target language $T$ that, given an unseen test document $x^T$ in $T$, predicts a binary (``Sick'' vs. ``Not Sick'') label. 

\begin{figure*}
    \centering
    \includegraphics[width=0.9\textwidth]{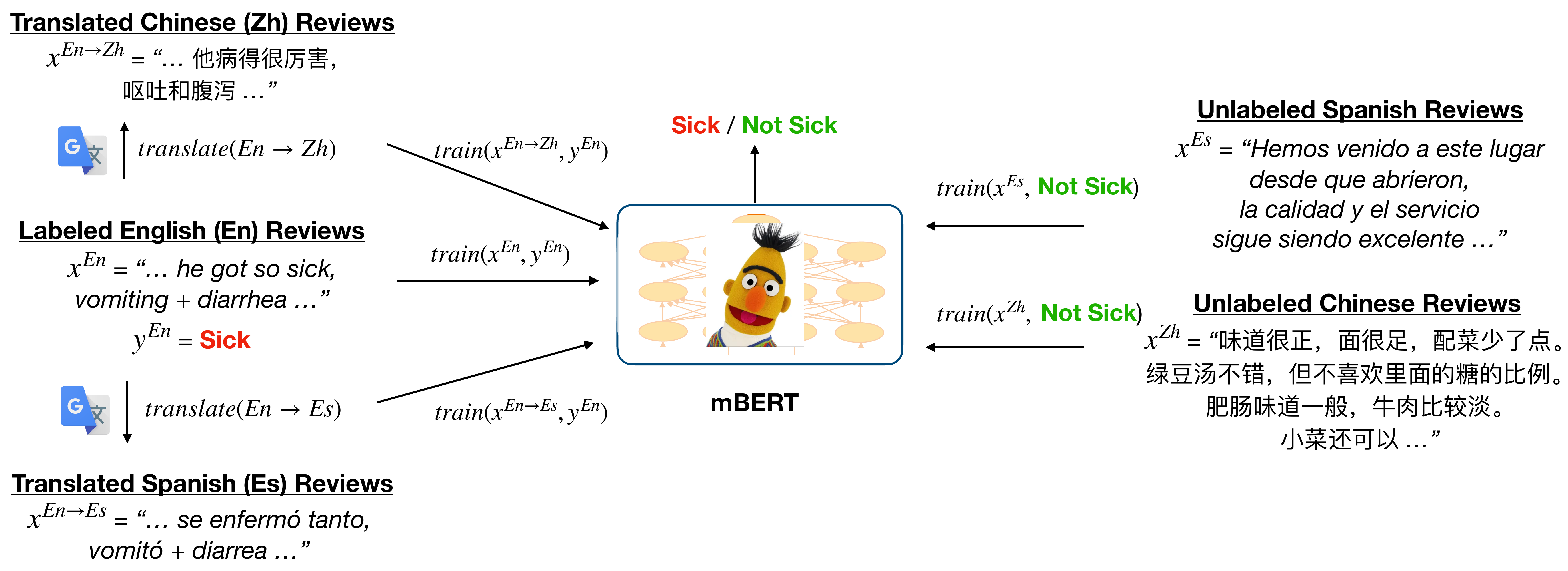}
    \caption{Our training procedure. We translate \emph{labeled} English reviews to the target languages and use the translated reviews with the original labels as extra training samples. We also use a sample of \emph{unlabeled} multilingual reviews as negative (``Not Sick'') training examples.}
    \label{fig:model_figure}
\end{figure*}

\subsection{Fine-Tuning mBERT on $S$ and $T$}
\label{s:model-finetuning-on-S-T}
To address the task mentioned in Section~\ref{s:problem-definition}, we use pre-trained mBERT representations, which effectively align representations of different languages (Section~\ref{s:background-crosslingual-text-classification}). 

It has been shown that mBERT achieves impressive zero-shot performance for tasks such as sentiment classification and topic detection~\cite{wu2019beto,pires2019multilingual}: fine-tuning mBERT on the labeled dataset $D_S$ in $S$ leads to accurate classification of unlabeled documents $x^T$ in $T$, possibly because representations across languages are well aligned with respect to the target sentiment or topic.
However, in contrast to previous tasks, we show that zero-shot mBERT is not effective for foodborne detection.
We hypothesize that this discrepancy is observed because pre-trained mBERT representations are not effectively aligned across languages with respect to the aspect of foodborne illness, which may be rarely mentioned in documents used for pre-training mBERT. 

To address this issue and improve classification performance for our task, we do not consider zero-shot training but fine-tune mBERT in both $S$ and $T$. %
Our main idea is that fine-tuning mBERT in documents from both $S$ and $T$ will encourage a stronger alignment of the cross-lingual representations with respect to the aspect of foodborne illness.  
The main challenge associated with our approach is that labeled documents are not available in the target language $T$. 

To generate training documents in $T$, we translate labeled documents $x_i^S$ from $S$ (English) to $T$ using machine translation. 
In particular, we assume that machine translation is sufficiently accurate to the extent that the translated document $x_i^{S\rightarrow T}$ has the same label as the original document $x_i^S$.
Under this assumption, we generate a weakly annotated dataset $D_T' = \{(x_i^{S \rightarrow T}, y_i^S)\}$ by translating all documents $x_i^S$ annotated as ``Sick'' and an equal number of documents randomly sampled from ``Not Sick'' documents in $D_S$.
Then, we increase the size of $D_T'$ by sampling unlabeled documents $x_i^T$ from $D_T$ uniformly at random. 
Each sampled document is assigned the ``Not Sick'' label as the chance of randomly choosing a document mentioning foodborne illness is very low. 
The number of sampled documents is chosen so that the total number of ``Not Sick'' documents in $D_T'$ is equal to that in $D_S$. 

After creating the weakly labeled $D_T'$ set we fine-tune our mBERT-based classifier jointly on $D_S$ and $D_T$ by concatenating and shuffling the two datasets. 
As we will show, this training procedure is more effective than fine-tuning mBERT on $D_S$ or $D_T$ separately. 

\subsection{Considering Multiple Source Languages}
\label{s:multiple-source-language}
Classification performance in $T$ may potentially improve using \emph{multiple} source languages $\{S_1, \dots, S_K\}$ other than $S$ (English) for which unlabeled documents and machine translation systems are available. 
The main idea behind this approach is that training signals from multiple source languages could prevent overfitting to a single source language and as a result encourage mBERT to learn better cross-lingual representations for our task.
Therefore, we adapt the procedure described in Section~\ref{s:model-finetuning-on-S-T} to consider more source languages in addition to $S$ and $T$, as we describe next. 
To train mBERT using multiple source languages $S, S_1, \dots, S_K$, we create a big training set that considers all source-language documents.
In particular, first we create a weakly-labeled dataset $D_{S_k}$ for each source language using machine translation, as we described in Section~\ref{s:model-finetuning-on-S-T} for creating $D_S$.
Then, we concatenate all source datasets $D_S, D_{S_1}', \dots, D_{S_K}'$ and fine-tune mBERT across all languages ($S, S_1, \dots, S_K, T$). 
Note that, in our preliminary experiments, we have treated all languages as equal but in the future it would be interesting to consider alternative approaches, such as using different weights for examples from different languages. 
Figure~\ref{fig:model_figure} shows our overall training procedure using English, Spanish and Chinese for training mBERT.

An important advantage of this approach is that the same mBERT classifier can be applied on any target language $T$ supported in mBERT. 
As a result, deployment in health departments would be easier since it involves a single model for all languages and does not require extra pre-processing steps such as running a language detector\footnote{In our experiments, language detectors sometimes predicted the wrong language for the text of a test restaurant review, for example because of multiple mentions of Italian dishes in a non-Italian review.} for each test document and applying language-specific models. 
Also, as we will show next, considering multiple source languages during training encourages better generalization to a new \emph{unseen} test language.

\section{Experiments}
\label{s:experiments}
We evaluate our approach on foodborne detection in English (En), Spanish (Es), Chinese (Zh), French (Fr), German (De), Japanese (Ja), and Italian (It).

\subsection{Experimental Settings}
\paragraph{Datasets.}
We use the same corpus of labeled English reviews from~\citet{effland2018discovering}.
This dataset contains English reviews with ground truth annotations provided by epidemiologists. 
Table~\ref{tab:english-dataset-stats} reports the number of reviews on the train and test set. 
For details, see~\citet{effland2018discovering}.

We collect unlabeled multilingual reviews from Yelp restaurants in New York City (NYC), Los Angeles (LA), as well as other metropolitan areas in the Yelp Challenge dataset.\footnote{\url{https://www.kaggle.com/yelp-dataset/yelp-dataset}} 
As the language of the reviews is not mentioned in the metadata, we used Python’s langdetect\footnote{\url{https://pypi.org/project/langdetect/}} library to automatically detect the language. 
For evaluation on non-English languages, we translate the 2975 English test reviews to the target languages using the Google Translate API.\footnote{The Google Translate API was used in February 2020.} 

\begin{table}[t]
\centering
\begin{tabular}{|l|c|c|c|}
\hline
 & \textbf{All Reviews} & \textbf{Sick} & \textbf{Not Sick}  \\ \hline
\textbf{Train} & 21,551 & 5894 & 15,657 \\ \hline
\textbf{Validation} & 1500 & 1090 & 410 \\ \hline
\textbf{Test} &  2975 & 949 & 2026 \\ \hline
\end{tabular}
\caption{Number of Yelp reviews in the English dataset with ground-truth (Sick vs. Not Sick) annotations.}
\label{tab:english-dataset-stats}
\end{table}

\begin{table}[t]
\centering
\resizebox{\columnwidth}{!}{
\begin{tabular}{|l|c|c|c||c|}
\hline
~                   & \textbf{NYC}  & \textbf{LA}    & \textbf{Yelp}  & \textbf{Total}  \\ 
~                   & \textbf{Area}  & \textbf{Area}    & \textbf{Challenge}  &   \\ 
\hline
\textbf{Spanish}             & 6267 & 11,458 & 2658  & 20,383  \\ 
\hline
\textbf{Chinese} & 1624 & 1488  & 603   & 3715   \\ 
\hline
\textbf{French}              & 3882 & 741   & 24,807 & 29,430  \\ 
\hline
\textbf{German}              & 2912 & 657   & 1394  & 4963   \\ 
\hline
\textbf{Japanese}            & 2161 & 1469  & 563   & 4193   \\ 
\hline
\textbf{Italian}             & 1259 & 322   & 173   & 1754   \\
\hline
\end{tabular}}
\caption{Number of unlabeled Yelp reviews from the New York City area, Los Angeles area, as well as other metropolitan areas in the Yelp Challenge dataset.}
\end{table}

\paragraph{Model Comparison.}
We compare the following models for our task:
\begin{itemize}
    \item \textbf{Monolingual LogReg}: the logistic regression classifier that achieved the best results in~\cite{effland2018discovering}. 
    We train LogReg for a non-English target language $T$ by translating English reviews to $T$ using Google Translate (see Section~\ref{s:model-finetuning-on-S-T}).
    \item \textbf{Monolingual BERT}: a monolingual BERT classifier. Similarly to LogReg, we train BERT for a non-English target language $T$ by translating English reviews to $T$ using Google Translate.
    \item \textbf{mBERT}: a multilingual BERT classifier. We train mBERT on several combinations of languages using our approach described in Section~\ref{s:foodborne-model}. 
\end{itemize}

\begin{table*}[t]
    \centering
    \begin{tabular}{|c|c|c|c|c|c|c|c|c||c|}
    \hline 
               &\textbf{Train} & \multicolumn{7}{c||}{\textbf{Test Language}}  & \textbf{AVG}\\ 

        \textbf{Model}     & \textbf{Language} & \textbf{En} & \textbf{Es} & \textbf{Zh} & \textbf{Fr} & \textbf{De} & \textbf{Ja} & \textbf{It} & \textbf{F1} \\\hline 
         Monolingual LogReg    & $T$ & 83.7 & 83.6 & 83.3 & 84.9 & 80.4 & 81.7 & 83.6 & 83.0\\
         Monolingual BERT     & $T$ & \textcolor{red}{\textbf{91.6}}  & \textcolor{red}{\textbf{91.3}} & \textbf{87.3} & \textcolor{red}{\textbf{92.4}} & \textbf{88.9} & \textbf{87.0} & \textcolor{red}{\textbf{90.7}} & \textcolor{red}{\textbf{89.9}}\\\hline 
        mBERT     & En & 89.0  & 82.0 & 78.8 & 80.6 & 59.0 & 65.5 & 67.5 & 74.6\\
        mBERT     & $T$ & 89.0  & 87.1 & 87.0 & 88.6 & 87.3 & 88.8 & \textbf{89.4} & 88.2\\
        mBERT     & En + $T$ & 89.0  & \textbf{89.8} & \textcolor{red}{\textbf{89.7}} & 90.5 & 88.0 & \textcolor{red}{\textbf{89.4}} & 88.2 & 89.2\\
        mBERT     & ALL& \textbf{91.3}  & 89.6 & 88.0& \textbf{90.7} & \textcolor{red}{\textbf{89.5}} & 86.8 & 89.2 & \textbf{89.3}\\
    \hline 
    \end{tabular}
    \caption{F1 scores for various approaches evaluated on different test languages.
    Monolingual LogReg and BERT are trained on the translated documents in the target language $T$. mBERT is trained with various language configurations. Training mBERT in English and $T$ is more effective than training on either language separately. Training mBERT across all 7 languages (``ALL'') leads to further improvements for En, Fr, and De. Results in red correspond to the best performance across all models. }
    \label{tab:main-results}
\end{table*}

\paragraph{Model Configuration.}
For LogReg, we tokenize text using Spacy\footnote{\url{https://spacy.io/api/tokenizer}} and convert the text documents to TF-IDF vectors. 

For monolingual BERT, we consider pre-trained BERT representations from huggingface\footnote{\url{https://huggingface.co}}:
\begin{itemize}
    \item English: \href{https://huggingface.co/bert-base-cased}{bert-base-uncased}
    \item Spanish: \href{https://huggingface.co/dccuchile/bert-base-spanish-wwm-cased}{dccuchile/bert-base-spanish-wwm-cased}
    \item Chinese:  \href{https://huggingface.co/bert-base-chinese}{bert-base-chinese}
    \item French: \href{https://huggingface.co/camembert-base}{camembert-base}
    \item German: \href{https://huggingface.co/bert-base-german-cased}{bert-base-german-cased}
    \item Japanese: \href{https://huggingface.co/cl-tohoku/bert-base-japanese}{cl-tohoku/bert-base-japanese}
    \item Italian: \href{https://huggingface.co/dbmdz/bert-base-italian-xxl-cased}{dbmdz/bert-base-italian-xxl-cased}
\end{itemize}

For mBERT, we consider pre-trained mBERT representations from huffingface: \href{https://huggingface.co/bert-base-multilingual-cased}{bert-base-multilingual-cased}.
We fine-tuned BERT and mBERT using the Python simpletransformers\footnote{\url{https://github.com/ThilinaRajapakse/simpletransformers}} library. 
We did a hyperparameter search with BERT on English data using the validation set. 
The best hyperparameters are a learning rate of 1e-05, a batch size of 512, and a maximum sequence length of 512. We fine-tune BERT/mBERT for up to 5 epochs with early stopping based on the validation loss. 

\paragraph{Evaluation Procedure.}
For each model, we choose the best set of hyperparameters according to the F1 score on the validation set. 
We report the following classification metrics on the test set: accuracy (Acc), precision (Prec), recall (Rec), and macro-average F1 score (F1). 

\begin{table}[t]
    \centering
    \begin{tabular}{|c|c|c|c|c|}
    \hline 
        \textbf{Model}     & \textbf{Train} & \textbf{Es} & \textbf{Zh} & \textbf{AVG}\\\hline 
        mBERT     & En  & 82.0 & 78.8 & 80.4\\
        mBERT     & ALL-$T$ & \textbf{84.7} & \textbf{84.0}& \textbf{84.4}\\
    \hline 
    \end{tabular}%
    \caption{Zero-shot performance under two different settings: training on English-only data (En) vs. training on all languages except the target language (ALL-$T$). The latter approach performs substantially better than the former.}
    \label{tab:zero_shot_performance}
\end{table}

\subsection{Experimental Results}
Table~\ref{tab:main-results} shows F1 scores on all languages for various methods.

\paragraph{Monolingual BERT outperforms previous systems.}
Monolingual BERT outperforms LogReg: leveraging pre-trained contextual representations captures foodborne illness effectively.

\paragraph{Monolingual BERT outperforms mBERT.}
Interestingly, monolingual BERT performs better than mBERT. We hypothesize that, by focusing on a single language, pre-trained monolingual BERT representations capture foodborne-related aspects more effectively than mBERT representations that were pre-trained for all languages in parallel.

\paragraph{Zero-shot mBERT is not effective.}
Training zero-shot mBERT using only English training data (En) is not effective and performs substantially worse than monolingual LogReg.
This result validates our argument that pre-trained mBERT representations do not effectively capture the aspect of food poisoning, which is rarely mentioned in documents used for pre-training mBERT.  

\begin{table}[t]
    \resizebox{\columnwidth}{!}{
    \begin{tabular}{|l|l||c|c|c|c|}
        \hline
        \textbf{Model}  & \textbf{Train}          & \textbf{Acc}        & \textbf{Prec}        & \textbf{Rec}            & \textbf{F1}             \\ \hhline{|b:=:=:=:=:=:=:b|}
        LogReg & En             & 88.1            & 74.1             & 96.2             & 83.7          \\ \hline
        BERT   & En             & \textbf{94.4}   & 88.1             & \textbf{95.4}             & \textbf{91.6} \\ \hline
        mBERT  & En             & 92.5            & 83.8             & 95.0             & 89.0          \\ 
        mBERT  & ALL            & 94.3           & \textbf{89.2}    & 93.6             & 91.3        \\ \hline 
    \end{tabular}}
    \caption{Evaluation on English Yelp reviews.}
    \label{tab:test-english}

\end{table}%
\begin{table*}[h!]
\centering

\begin{subtable}{\columnwidth}
    \resizebox{\columnwidth}{!}{
    \begin{tabular}{|l|l|l|l|l|l|}
        \hline
        Model  & Train         & Acc         & Prec         & Rec            & F1             \\ \hhline{|b:=:=:=:=:=:=:b|}
        LogReg & Es            & 87.9            & 73.7             & \textbf{96.4}    & 83.6          \\ 
        LogReg* & En           & 88.2            & 75.5             & 93.4             & 83.5          \\ \hline
        BERT   & Es            & \textbf{94.2}   & 87.1             & 96.0             & \textbf{91.3} \\ 
        BERT*   & En           & 93.6            & 87.1             & 93.9             & 90.4          \\ \hline
        mBERT  & En            & 89.7            & \textbf{92.3}             & 73.8             & 82.0          \\ 
        mBERT  & Es            & 90.9            & 79.5             & \textbf{96.4}    & 87.1          \\ 
        mBERT  & En+Es         & 93.2            & 85.9             & 94.1             & 89.8          \\ 
        mBERT  & ALL           & 93.3            & 89.0             & 90.2             & 89.6          \\ \hline
    \end{tabular}}
    \caption{Results on Spanish.}
\end{subtable}%
\quad
\begin{subtable}{\columnwidth}
    \resizebox{\columnwidth}{!}{
    \begin{tabular}{|l|l|l|l|l|l|l|}
        \hline
        Model  & Train     & Acc         & Prec         & Rec            & F1             \\ \hhline{|b:=:=:=:=:=:=:b|}
        LogReg &Zh         & 88.3            & 76.8             & 90.9             & 83.3          \\ 
        LogReg* & En       & 87.2            & 76.9             & 85.6             & 81.0          \\ \hline
        BERT   & Zh        & 91.3            & 81.2             & 94.5             & 87.3          \\ 
        BERT*   & En       & 92.4            & 88.6             & 87.6             & 88.1          \\ \hline
        mBERT  & En        & 88.2            & \textbf{91.9}    & 69.0             & 78.8          \\ 
        mBERT  & Zh        & 90.9            & 80.2             & 95.0             & 87.0          \\
        mBERT  & En+Zh     & \textbf{93.2}   & 86.2             & 93.6             & \textbf{89.7} \\ 
        mBERT  & ALL       & 91.7            & 81.7             & \textbf{95.5}    & 88.0          \\ \hline
    \end{tabular}}
    \caption{Results on Chinese.}
\end{subtable}%

\begin{subtable}{\columnwidth}
    \resizebox{\columnwidth}{!}{
    \begin{tabular}{|l|l|l|l|l|l|}
        \hline
        Model  & Train         & Acc         & Prec         & Rec            & F1             \\ \hhline{|b:=:=:=:=:=:=:b|}
        LogReg & Fr            & 89.4            & 77.8             & 93.6             & 84.9          \\ \hline
        BERT   & Fr            & \textbf{95.0}   & 89.6             & \textbf{95.4}    & \textbf{92.4} \\ \hline
        mBERT  & En            & 88.9            & \textbf{91.4}    & 72.1             & 80.6          \\ 
        mBERT  & Fr            & 92.1            & 82.6             & \textbf{95.4}    & 88.6          \\ 
        mBERT  & En+Fr         & 93.6            & 86.4             & 94.9             & 90.5          \\ 
        mBERT  & ALL           & 94.0            & 89.8             & 91.6             & 90.7          \\ \hline
    \end{tabular}}
    \caption{Results on French.}
\end{subtable}%
\quad
\begin{subtable}{\columnwidth}
    \resizebox{\columnwidth}{!}{
    \begin{tabular}{|l|l|l|l|l|l|}
        \hline
        Model  & Train         & Acc         & Prec         & Rec            & F1             \\ \hhline{|b:=:=:=:=:=:=:b|}
        LogReg & De            & 85.2            & 69.5             & 95.1             & 80.4          \\ \hline
        BERT   & De            & 92.4            & 83.3             & \textbf{95.5}    & 88.9          \\ \hline
        mBERT  & En            & 81.2            & \textbf{97.1}    & 42.4             & 59.0          \\ 
        mBERT  & De            & 91.1            & 80.4             & 95.4             & 87.3          \\ 
        mBERT  & En+De         & 91.9            & 82.9             & 93.9             & 88.0          \\ 
        mBERT  & ALL           & \textbf{93.0}   & 86.0             & 93.4             & \textbf{89.5} \\ \hline
    \end{tabular}}
    \caption{Results on German.}
\end{subtable}%

\begin{subtable}{\columnwidth}
    \resizebox{\columnwidth}{!}{
    \begin{tabular}{|l|l|l|l|l|l|}
        \hline
        Model  & Train      & Acc         & Prec         & Rec            & F1             \\ \hhline{|b:=:=:=:=:=:=:b|}
        LogReg & Ja              & 86.5            & 71.9             & 94.7             & 81.7          \\ \hline
        BERT   & Ja              & 91.2            & 82.3             & 92.3             & 87.0          \\ \hline
        mBERT  & En              & 83.0            & \textbf{93.2}    & 50.5             & 65.5          \\ 
        mBERT  & Ja              & 92.4            & 83.6             & 94.7             & 88.8          \\ 
        mBERT  & En+Ja           & \textbf{92.8}   & 84.2             & 95.3             & \textbf{89.4} \\ 
        mBERT  & ALL             & 90.7            & 79.4             & \textbf{95.7}    & 86.8          \\ \hline
    \end{tabular}}
    \caption{Results on Japanese.}
\end{subtable}%
\quad
\begin{subtable}{\columnwidth}
    \resizebox{\columnwidth}{!}{
    \begin{tabular}{|l|l|l|l|l|l|}
        \hline
        Model  & Train      & Acc         & Prec         & Recall            & F1             \\ \hhline{|b:=:=:=:=:=:=:b|}
        LogReg & It              & 88.5            & 76.8             & 91.7             & 83.6          \\ \hline
        BERT   & It              & \textbf{93.7}   & 85.5             & 96.5             & \textbf{90.7} \\ \hline
        mBERT  & En              & 83.5            & \textbf{91.1}    & 53.6             & 67.5          \\ 
        mBERT  & It              & 92.8            & 84.6             & 94.8             & 89.4          \\
        mBERT  & En+It           & 91.7            & 81.1             & \textbf{96.6}    & 88.2          \\
        mBERT  & ALL             & 92.7            & 84.3             & 94.6             & 89.2          \\ \hline
    \end{tabular}}
    \caption{Results on Italian.}
\end{subtable}%
\caption{Results on different target languages. LogReg and BERT are trained on the translated target-language documents.  LogReg* and BERT* are trained on English and applied on test reviews by translating the corresponding text from the target language to English. mBERT is trained with various configurations.}
    \label{fig:detailed-results}
\end{table*}

\paragraph{Artificial training reviews in $T$ improve mBERT's performance.}
Translating English reviews to $T$ and using translated reviews to train mBERT on $T$ is substantially better than zero-shot mBERT trained on English directly.
This result highlights the importance of in-language training documents, even if those documents are artificially created. 
Furthermore, training mBERT jointly on English and the target language $T$ leads to better performance compared to training on each language separately. 

\paragraph{Training on all languages leads to the best performance for mBERT.}
On average across languages, mBERT trained on all languages jointly performs better than other mBERT configurations with a single source language, but comparably to mBERT trained on En and $T$.
Interestingly, for Chinese (Zh) and Japanese (Ja) performance is worse if more languages are added to the training set, possibly because these languages are more distant from Romance languages such as Spanish or French, and as a result considering those languages in the training set is not helpful.  

\paragraph{Using multiple source languages leads to higher zero-shot performance.}
Table~\ref{tab:zero_shot_performance} shows results for the setting where we assume that documents from the target language are not available for training. 
Crucially, training mBERT on all languages except this target language performs substantially better than training mBERT only on English data, validating the importance of training mBERT on multiple languages jointly. 
Also, F1 scores when ignoring those languages during training (ALL-$T$) are lower by about 5 absolute points compared to considering them during training (ALL): we could potentially apply our approach to any unseen language out of the 104 languages that are supported by mBERT.  

\paragraph{Detailed English results.}
Table~\ref{tab:test-english} shows results in English. 
BERT (monolingual) has the best F1 score. 
Training mBERT on all languages (En, Es, Zh, Fr, De, Ja, It) is more effective than training mBERT on English-only labeled data. 
This validates our hypothesis that, by considering all languages, mBERT generalizes better to test reviews. 

\paragraph{Detailed non-English results.}
Table~\ref{fig:detailed-results} shows detailed results on non-English datasets. 
For Spanish and Chinese we evaluated an additional baseline where test reviews are translated to English and considered by LogReg (``Logreg*'' baseline) or BERT (``BERT*'' baseline) that were trained on English reviews only. 
This approach is less effective, as well as more expensive than the other approaches: to deploy in health departments, it would require each new test review to be translated to English. 
While BERT has the highest F1 score on average over all approaches, mBERT has higher recall than BERT on most non-English target languages.
%

%

%
%

\paragraph{We detect reviews mentioning foodborne illness.}
To demonstrate the potential of our approach for detecting foodborne illness, we ran mBERT on unlabeled restaurant reviews from the NYC Area, LA Area, and the Yelp Challenge dataset. 
Table~\ref{tab:examples-unlabeled-sick-reviews} shows two examples that were classified as ``Sick'' by our classifier. 
Translating those two reviews to English and applying LogReg (trained in English) led to a (wrong) ``Not Sick'' prediction, possibly because the translated reviews are not matching the training distribution for LogReg. 

\begin{CJK*}{UTF8}{gbsn}
\begin{table*}[t]
\centering
\begin{tabular}{|p{1.5cm}|p{14.5cm}|}
\hline
\multirow{2}{*}{\textbf{Spanish}}  & \multicolumn{1}{p{12.5cm}|}{\textbf{Original (Es) text:} Definitivamente mi peor experiencia, me intoxique con un ostra mala, llevo 4 días en muy malas condiciones, por favor tengan cuidado, los ostiones y mariscos no se pueden comer en cualquier lugar, yo aprendi por las malas, espero que mi experiencia le sirva a alguien}\\ 
& 
\multicolumn{1}{p{12.5cm}|}{\textbf{mBERT (train: ALL) prediction:} \textcolor{blue}{\textbf{``Sick''} \cmark}} \\
\cline{2-2} & \multicolumn{1}{p{12.5cm}|}{\textbf{Translated (En) text:} Definitely my worst experience, I got intoxicated with a bad oyster, I have been in very bad conditions for 4 days, please be careful, the oysters and shellfish cannot be eaten anywhere, I learned through the bad ones, I hope my experience will serve you someone}\\
& 
\multicolumn{1}{p{12.5cm}|}{\textbf{LogReg (train: En) prediction:} \textcolor{red}{\textbf{``Not Sick''} \xmark }}\\
\hline\hline

\multirow{2}{*}{\textbf{Chinese}}  & \multicolumn{1}{p{12.5cm}|}{\textbf{Original (Zh) text:} 装修和服务都还不错，但味道极差：底料没有味道，我们自己加了几次盐和料才勉强能吃，菜品也非常不新鲜。一顿饭吃得我们四个人都很生气，然后回家三个人都拉肚子。绝对不会再去吃。Avoid!!
}\\ 
& 
\multicolumn{1}{p{12.5cm}|}{\textbf{mBERT (train: ALL) prediction:} \textcolor{blue}{\textbf{``Sick''} \cmark}} \\
\cline{2-2} & \multicolumn{1}{p{12.5cm}|}{\textbf{Translated (En) text:} The decoration and service are good, but the taste is very bad: the base material has no taste, we added salt several times to make it barely edible, and the dishes are very fresh. Four of us were angry at a meal, and then all three got diarrhea. Will never eat again. Avoid !!}\\
& 
\multicolumn{1}{p{12.5cm}|}{\textbf{LogReg (train: En) prediction:} \textcolor{red}{\textbf{``Not Sick''} \xmark }}\\
\hline\hline

\multirow{2}{*}{\textbf{German}}   & \multicolumn{1}{p{12.5cm}|}{\textbf{Original (De) text:} Wir haben hier 2 bowls mit Steak und einen Burger gegessen. Für unverschämte 70,03\$ gab es recht kleine und nicht wirklich gute Portionen (besonders die bowls). Nachdem mein Sohn von der Bowl gegessen hat, musste er brechen. Auch meiner Tochter und mir war schlecht. Der Service wirkte lieblos und desinteressiert. Die bowls kamen gerade mal lauwarm an unseren Tisch und die Chips vom Burger schmeckten nach nichts. Nicht zu empfehlen!!!} \\ 
&
\multicolumn{1}{p{12.5cm}|}{\textbf{mBERT (train: ALL) prediction:} \textcolor{blue}{\textbf{``Sick''} \cmark}} \\
\cline{2-2} 
                          & \multicolumn{1}{p{12.5cm}|}{\textbf{Translated (En) text:} We ate 2 bowls of steak and a burger here. For outrageous \$70.03 there were quite small and not really good portions (especially the bowls). After my son ate from the bowl, he had to break. My daughter and I were also bad. The service seemed careless and uninterested. The bowls just came to our table lukewarm and the chips from the burger didn't taste like anything. Not recommendable!!!} \\ 
                         & \multicolumn{1}{p{12.5cm}|}{\textbf{LogReg (train: En) prediction:} \textcolor{red}{\textbf{``Not Sick''} \xmark }}\\

                          \hline
\end{tabular}
\caption{Examples of Spanish, Chinese and German restaurant reviews in our dataset classified as ``Sick'' and their translations to English.}
\label{tab:examples-unlabeled-sick-reviews}
\end{table*}
\end{CJK*}

\section{Discussion and Future Work}
\label{s:conclusion}
We presented our cross-lingual learning method for scaling foodborne illness detection to languages beyond English without extra annotations for non-English languages.
As most reviews do not discuss foodborne illness, it is challenging to create proper evaluation datasets for all languages. 

In our preliminary experiments, we evaluated our approach on non-English languages by translating labeled test reviews from English to other languages. 
A caveat of this evaluation approach is that complaints of foodborne illness in native-language reviews may be expressed differently than in automatically translated reviews and thus, performance numbers may not be fully indicative of performance in native reviews.
Therefore, an important next step is to create better evaluation datasets.

Our exploratory results show that training mBERT in multiple languages jointly is more effective than training mBERT on English (zero-shot approach) or the target-language only. 
On average across languages mBERT is outperformed by monolingual BERT trained on (translated) target-language documents. 
On the other hand, deploying mBERT in health departments for daily inspections would be easier as it would not require extra pre-processing steps such as language detection that may introduce errors. 
Also, we showed that mBERT could potentially be applied for languages that were not seen in the training set, without extra translation efforts.   

As another interesting direction for future work, we plan to evaluate the cross-lingual transfer approach of~\citet{karamanolakis2019clts}, which applies even for low-resource languages that are not supported by mBERT or for which machine translation systems are not available.
We also plan to extend our system for predicting which languages to use as source languages to achieve good performance on a target language~\cite{lin2019choosing}. 

\subsubsection*{Acknowledgments}
We thank the anonymous reviewers for their constructive feedback. This material is based upon work supported by the National Science Foundation under Grant No. IIS-15-63785.

\bibliographystyle{acl_natbib}
\bibliography{anthology,emnlp2020}
\end{document}